\documentclass[10pt, a4paper]{article}

\newcommand{\corpusname}[0]{\textsc{EvoGrad}}

\usepackage{lrec-coling2024} 

\usepackage{gb4e}
\usepackage{hyperref}

\usepackage[utf8]{inputenc}

\usepackage{amsmath}
\usepackage{amssymb}
\usepackage{pifont}
\usepackage{breqn}
\usepackage{multirow}
\usepackage{booktabs}
\usepackage{xcolor}
\usepackage{tikz-qtree}
\usepackage{float}
\usepackage{tablefootnote}

\newcolumntype{L}[1]{>{\raggedright\let\newline\\\arraybackslash\hspace{0pt}}m{#1}}
\newcolumntype{C}[1]{>{\centering\let\newline\\\arraybackslash\hspace{0pt}}m{#1}}
\newcolumntype{R}[1]{>{\raggedleft\let\newline\\\arraybackslash\hspace{0pt}}m{#1}}

\title{\corpusname: A Dynamic Take on the Winograd Schema Challenge with Human Adversaries}

\name{Jing Han Sun, Ali Emami} 

\address{University of Montreal/Mila, Brock University \\
         Montreal, Canada, Saint Catharines, Canada \\
         jing.han.sun@umontreal.ca, aemami@brocku.ca\\
         }

\abstract{
While Large Language Models (LLMs) excel at the Winograd Schema Challenge (WSC), a coreference resolution task testing common-sense reasoning through pronoun disambiguation, they struggle with instances that feature minor alterations or rewording. To address this, we introduce \corpusname{}, an open-source platform that harnesses a human-in-the-loop approach to create a dynamic dataset tailored to such altered WSC instances. Leveraging ChatGPT's capabilities, we expand our task instances from 182 to 3,691, setting a new benchmark for diverse common-sense reasoning datasets. Additionally, we introduce the \textit{error depth} metric, assessing model stability in dynamic tasks. Our results emphasize the challenge posed by \corpusname{}: Even the best performing LLM, GPT-3.5, achieves an accuracy of 65.0\% with an average error depth of 7.2, a stark contrast to human performance of 92. 8\% accuracy without perturbation errors. This highlights ongoing model limitations and the value of dynamic datasets in uncovering them.
 \\ \newline  }

\begin{document}

\maketitleabstract

\section{Introduction}

The Winograd Schema Challenge (WSC), a co-reference resolution task, was developed to gauge the common-sense reasoning of automated systems \cite{winograd1972, levesque2011winograd}.  Given subtly varying sentence pairs, the task is to correctly associate a pronoun with a noun, as illustrated below:

\begin{exe}
\ex\begin{xlist}
\ex\label{ex-1a}
Tom told Ralph, ``Check,'' as he \underline{moved} \textit{his} bishop. (Answer: Tom)
\ex\label{ex-2a} Tom told Ralph, ``Check,'' as he \underline{took} \textit{his} bishop. (Answer: Ralph)
\end{xlist}
\end{exe}

In these examples, chess knowledge informs our interpretation of the pronoun \textit{his}—either referring to \textit{Tom} or \textit{Ralph}—based on the action performed, either a \textit{move} or \textit{take}. While humans find such tasks intuitive, they pose a challenge for statistical models, especially when lacking exposure to basic rules or common knowledge. Yet, recent developments of extensive common-sense reasoning datasets and benchmarks have allowed LLMs to achieve near-human performance on WSC variants \cite{brown2020language,sakaguchi2020winogrande}. This impressive accomplishment raises the question: has the WSC, seen as a definitive alternative to the Turing Test, been definitively ``defeated'' \cite{kocijan2022defeat}?

\begin{figure*}[ht]
\begin{center}
\includegraphics[width=\textwidth]
{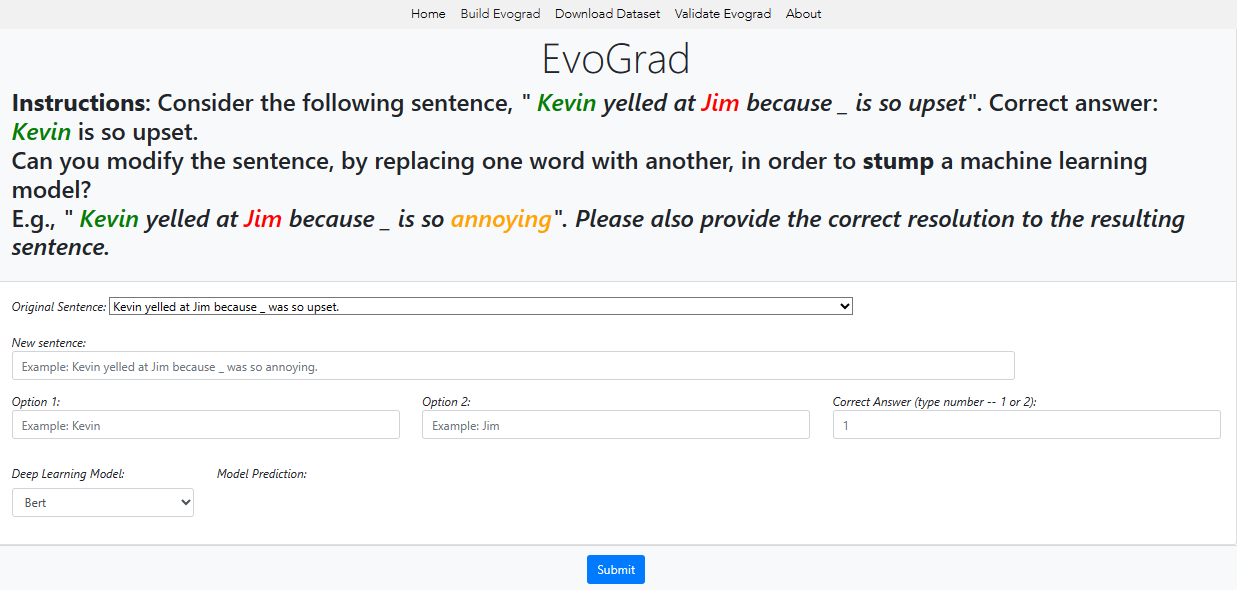}
\caption{Interface of \corpusname{} at \url{https://evograd.com}}
\label{fig:interface}
\end{center}
\end{figure*}

At the same time, evidence suggests that even slight alterations to a WSC task can significantly undermine a model's performance \cite{jia-liang-2017-adversarial, trichelair2018on, trichelair-etal-2019-reasonable, balasubramanian-etal-2020-whats, lin-etal-2020-birds,elazar-etal-2021-back}. This instability may reflect a discrepancy between current supervision paradigms and the dynamic nature of common sense acquisition. It suggests the potential value of exploring various approaches, including the human-and-model-in-the-loop concept, as part of a broader strategy to address these challenges \cite{nie-etal-2020-adversarial, kiela-etal-2021-dynabench, lu-etal-2022-rationale}.

Existing datasets, often curated by select scientific communities or crowdsourcing platforms, may also unintentionally bias models toward certain knowledge instances or values, which may not be universally shared. This consideration underscores the need for diverse, dynamic, and inclusive benchmarks in the journey towards systems equipped with generalized common sense.

Consider the chess example mentioned earlier. While the original WSC sentences test the model's understanding of the game's basic rules, perturbations can further probe deeper nuances and potential biases:

\begin{exe}
\ex\begin{xlist}
\ex\label{ex-1b}
Maria told Jane, ``Your move,'' as she \underline{adjusted} \textit{her} queen. (Answer: Maria) \ex\label{ex-2b} Maria told Jane, ``Your move,'' as she \underline{glanced} at \textit{her} clock. (Answer: Jane)
\end{xlist}
\end{exe}

In these variations, the emphasis shifts from the action performed on a chess piece to the broader context of a timed chess match. Slight word changes can dramatically alter the correct answer, exposing potential model biases or gaps in understanding. Such perturbations, especially when generated by diverse human contributors, ensure a broader and more comprehensive test of a model's common-sense reasoning capabilities.

In this paper, we propose a revisit to the WSC within the framework of human-and-model-in-the-loop. We introduce 
\corpusname, an open-source, user-centric platform dedicated to the active generation and expansion of nuanced examples for the WSC through human-in-the-loop interactions. Our work contributes three primary advancements:

\textbf{A novel data construction mechanism:} We enhance the WSC with our unique approach to human-adversarial perturbations, combining human creativity with the efficiency of ChatGPT. This innovative union, along with our use of Wordnet for synonym-based variation, led to a dataset expansion from 182 to 3691 instances, setting a new standard for dynamic, diverse, and high-quality common-sense reasoning datasets. Notably, our evaluations highlight the challenging nature of \corpusname{}, revealing significant gaps in model abilities when compared to human benchmarks.

\textbf{A new metric for model stability:} In response to the instability of transformer-based models on WSC-like tasks \cite{abdou-etal-2020-sensitivity}, we introduce a metric termed \textit{error depth}. This measure, derived from our data construction process, offers a quantifiable assessment of model stability. We advocate for its inclusion in evaluation reports alongside accuracy-based metrics, which could discourage the development of models that achieve high scores due to incorrect reasoning.

\textbf{Online platform for user contributions:} Available at \url{https://evograd.com}\footnote{All aspects of the website remain anonymous during the submission and review process to maintain the integrity of the user-contributed data and ensure unbiased evaluation.}, our platform encourages public participation in the continuous expansion of the dataset. Users can modify existing task instances and observe the predictions of a chosen LLM, fostering a more participatory and immersive data construction process (Figure \ref{fig:interface}).

\begin{figure*}[ht]
    \begin{tikzpicture}[level distance=60pt]
    \tikzset{every tree node/.style={text width=13em,text centered}}
        \Tree [.{$s_0$: Kevin yelled at Jim because he was so upset.} 
                \edge node[auto=right]{\small{$per_2(s_0,\text{screamed})$}}; 
                    [.{$s_{1(2)}$: Kevin \textcolor{cyan}{screamed} at Jim because he was so upset.} 
                        \edge node[auto=right]{\small{$per_4(s_{1(2)},\text{Melissa})$}};
                        [.{$s_{2(2,4)}$: Kevin \textcolor{cyan}{screamed} at \textcolor{orange}{Melissa} because he/she was so upset.} ] 
                        \edge node[auto=left]{\small{$per_5(s_{1(2)},\text{although})$}};
                        [.{$s_{2(2,5)}$: Kevin \textcolor{cyan}{screamed} at Jim \textcolor{orange}{although} he was so upset.} 
                        ] ] 
                \edge node[auto=left]{\small{$per_9(s_0,\text{annoying})$}}; 
                    [.{$s_{1(9)}$: Kevin yelled at Jim because he was so \textcolor{cyan}{annoying}.} 
                        \edge node[auto=left]{\small{$per_7(s_{1(9)},\text{became})$}};
                        [.{$s_{2(7,9)}$:  Kevin yelled at Jim because he \textcolor{orange}{became} so \textcolor{cyan}{annoying}.} ] ] ]
    \end{tikzpicture}
    \caption{Evolution figure of the sentence ``Kevin yelled at Jim because he was so upset.'' up to depth level 2.}
    \label{fig:evolution_example}
\end{figure*}

\section{Related Work}

\subsection{WSC-based Datasets}
\label{sec:wsc_datasets}
\looseness=-1 
The Winograd Schema Challenge (WSC) \cite{levesque2011winograd} inspired various datasets for pronominal coreference resolution, each tackling specific challenges in the WSC or model evaluations. Datasets like Winogrande \cite{sakaguchi2020winogrande} and KnowRef \cite{emami-etal-2019-knowref} address the WSC's size constraints. WinoGender \cite{rudinger-etal-2018-gender}, WinoBias \cite{zhao-etal-2018-gender}, and KnowRef-60k \cite{emami2020analysis} focus on model biases, while WinoWhy \cite{zhang-etal-2020-winowhy} and WinoLogic \cite{he-etal-2021-winologic} target common sense deficiencies in models. Some research efforts enhanced the original WSC task \cite{wang-etal-2018-glue, trichelair2018on, kocijan-etal-2019-surprisingly, elazar-etal-2021-back, zahraei2024wsc+} and utilized crowd-sourcing for task development \cite{isaak2019winoflexi, sakaguchi2020winogrande}. While these static datasets each offer distinct strengths, they often introduce challenges that necessitate prolonged research and iterations. EvoGrad, on the other hand, adopts a dynamic framework, allowing for swift adjustments and refinements in response to emerging challenges.

\subsection{Dynamic Datasets}

Dynamic datasets, updated over time to present new challenges, have been developed for various tasks \cite{zellers-etal-2019-hellaswag, lin-etal-2020-commongen}. Adversarial frameworks, as seen in Adversarial SQuAD, SWAG, HellaSWAG, CODAH and ANLI, exemplify this approach \cite{jia-liang-2017-adversarial, zellers-etal-2018-swag, zellers-etal-2019-hellaswag, chen-etal-2019-codah, nie-etal-2020-adversarial}. Techniques such as AFLite address biases through adversarial filtering \cite{lebras2020adversarial}, while other methods use continuous learning or a human-model collaborative process \cite{lan-etal-2017-continuously, yang-etal-2018-hotpotqa, wallace-etal-2019-trick, dinan-etal-2019-build, nie-etal-2020-adversarial, xu-etal-2021-bot, kiela-etal-2021-dynabench}. ANLI and Dynabench are notable for their multi-round adversarial data collection \cite{nie-etal-2020-adversarial, kiela-etal-2021-dynabench}. \corpusname{}, while aligning with the dynamic dataset philosophy, specifically targets WSC-based tasks. It merges human-and-model collaboration, continuous learning, and domain-specific insights for evolutionary data creation, amplifying the depth and relevance of WSC challenges to shed light on common-sense reasoning.

\subsection{Data Augmentation Methods in NLP}

Data augmentation techniques in NLP create new examples from existing ones, obviating the need for novel data collection \cite{shi-etal-2021-substructure, feng-etal-2021-survey}. These methods include token-level manipulation, text generation restricted, soft data enhancement, and structure-aware data augmentation \cite{wang-yang-2015-thats, bergmanis-etal-2017-training, zhang2018mixup, xu-etal-2016-improved}. Our approach, mainly a token-level manipulation technique, extends beyond the substitution of words to include the addition and removal of tokens, allowing more significant sentence transformations \cite{zmigrod-etal-2019-counterfactual, lu2020gender, shi-etal-2018-learning}. We also measure the \textit{depth} of changes (Section \ref{sec:metric}) relative to the original sentence, providing insights into model stability as a function of perturbations.

\subsection{Large Language Models in Data Augmentation and Annotation}
Large language models have emerged as effective tools for NLP data augmentation and annotation, often exceeding the performance of crowd-workers in terms of efficiency and cost \cite{gilardi2023chatgpt}. These models have been shown to be effective in tasks such as zero-shot gender identification and providing explanations for implicit hate speech \cite{kuzman2023chatgpt, huang2023chatgpt}. AugGPT, for instance, outperforms traditional text augmentation methods in few-shot learning scenarios by rephrasing sentences \cite{dai2023chataug}. Similarly, ChatGPT has shown potential to simplify social computing tasks by replicating human-like annotations \cite{zhu2023can}. Building on these insights, we introduce an enhanced data augmentation method that encompasses token substitutions, additions, and removals, aiming to address common-sense reasoning deficiencies in the WSC and related tasks.

\section{\corpusname{}}

\subsection{Dataset Evolution by Perturbation}
\label{sec:dataset_construction}
\looseness=-1 We adopt an evolutionary approach to dataset expansion, initiating the process with randomly selected instances from the original Winograd Schema Challenge (WSC273) \cite{levesque2011winograd} and Winogrande \cite{sakaguchi2020winogrande}, which are correctly resolved by all evaluated models.

Our method introduces a one-word perturbation to each sentence, effectively mutating it via substitution. We define a perturbation function $per_j(s,w)$ that replaces the token at index $j$ in sentence $s$ with the token $w$. Though primarily substitution-based, this function can also facilitate the addition or removal of words, denoted as $per_j(s,w_j + w)$ and $per_j(s,\epsilon)$ respectively, with $\epsilon$ symbolizing an empty string.

The function is generalized as follows:

\vspace{-6mm}

\begin{multline}
\label{perturbation_equation}
    per_j(s_{k(i_1,...,i_k)}, w) = s_{(k+1)(i_1,...,i_j,...,i_{k+1})} \\
    j \not\in \{i_1,...,i_k\} \text{ \& } i_1 < ... < i_{k+1}
\end{multline}

In this equation, $s_{k(i_1,...,i_k)}$ signifies the $k$th perturbation on the base sentence $s_0$, wherein tokens at indices $i_1,...,i_k$ have been modified from $s_0$ (Equation \ref{perturbation_equation}). The term $k$ denotes the `depth' or generation of the sentence.

The conditions set for $j$ and indices $i_1,...,i_{k+1}$ ensure that a depth increment corresponds solely to the perturbation of a token distinct from those previously perturbed (i.e.,${i_1,...,i_k}$). Although repeated modifications at the same token position are not prohibited, such sentences maintain their original depths. This approach follows our depth interpretation, emphasizing model stability against sentences that are increasingly divergent from the original. This methodological choice facilitates the systematic generation of progressively varied sentences, thereby enriching the dataset.

The perturbation function is applied iteratively, generating a cascade of output instances from each input instance. This process is illustrated in Figure \ref{fig:evolution_example} by the sentence `Kevin yelled at Jim because he was so upset.' Through several iterations of the perturbation function, we generate a wide spectrum of sentences, each incrementally divergent from the original.

\subsection{Scaling with ChatGPT}
\label{sec:scaling_chatgpt}

Beyond user contributions, we strategically employed ChatGPT\footnote{https://chat.openai.com/} to vastly expand our dataset. We initialized the process with 14 seed sentences (7 from WSC273 and 7 from Winogrande-valid) and designed an elaborate prompt that enabled ChatGPT to act as an `expert human annotator'. The prompts were meticulously crafted to guide the model generation process via demonstrative examples and called for frequent self-reflection to ensure the quality of the output. One unique aspect of these prompts was the incorporation of a segmented generation process, interspersed with feedback to ensure quality control and continuous self-assessment. For each instance, we verified semantic coherence and implemented a validation step to ensure pronouns and co-references matched commonly accepted or typical human readings. An illustrative dialogue sample can be found in the Appendix in section \ref{sec:chatgpt_dialogue}.

This rigorous approach to prompt engineering culminated in the generation of approximately 100 new instances per seed sentence. We further diversified these generated sentences by modifying words, altering the correct antecedent, and varying the total perturbation depth from the original sentences. This strategy effectively harnessed the power of human creativity and the scalability of the model to significantly expand our dataset. As a result, we managed to augment our initial 182-instance dataset to a much more extensive collection of 1,414 sentences, thereby facilitating a more comprehensive evaluation of model performance on dynamic WSC tasks.

\subsection{Scaling with Wordnet}
\label{sec:wordnet}
To increase the diversity of our dataset, we utilized Wordnet \cite{fellbaum2010wordnet}, a lexical database, to augment the 1,414 sentences obtained from our \textit{ChatGPT Scaling} stage. This process enabled us to nearly triple our dataset size to a final count of 3,691 sentences.

Our strategy was to introduce variability while preserving the context of the sentence and grammatical accuracy. We achieved this by iterating over each sentence and randomly selecting a word—excluding stop words and named entities—for replacement. Once a word was selected, a random synonym from Wordnet was chosen as its substitute.  In cases where the chosen word was a verb, we ensured that the replacement synonym matched the tense of the original verb.

This approach allowed us to maintain the integrity of our original dataset while significantly enhancing its size and complexity. The resulting sentences provided a rich basis for model testing, aiding in the generation of a more diverse and nuanced set of pronoun disambiguation scenarios.

\begin{table}[ht]
\centering
\begin{tabular}{p{2cm} p{0.5cm} p{0.5cm} p{3cm}}
\toprule
\textbf{Dataset} & \textbf{Sub} & \textbf{Size} & \textbf{Method} \\ 
\midrule
\corpusname{}-S & - & 182 & Human (14 orig.) \\
\addlinespace
\corpusname{}-M & Train & 1010 & ChatGPT (1-10) \\
 & Val & 202 & ChatGPT (11-12) \\
 & Test & 202 & ChatGPT (13-14) \\
\addlinespace
\corpusname{}-L & Train & 2963 & WordNet (M Train) \\
 & Val & 526 & WordNet (M Val) \\
 & Test & 202 & ChatGPT (13-14) \\
\bottomrule
\end{tabular}
\caption{Summary of \corpusname{} Allocation}
\label{tab:dataset}
\end{table}

\begin{table*}[h]
\centering
\begin{tabular}{>{\raggedright\arraybackslash}p{2.5cm}p{8cm}l>{\centering\arraybackslash}p{1cm}}
\toprule
\textbf{Source} & \textbf{Sentence} & \textbf{Answer} & \textbf{Depth} \\ 
\midrule
Original (WSC) & I poured water from the bottle into the cup until \_ was full. & cup & 0 \\ 
\addlinespace
Human-perturbed & I poured water from the bottle into the cup because \_ was empty. & cup & 2 \\ 
\addlinespace
ChatGPT-scaled & I poured water from the bottle, filling the cup until \_ was empty. & bottle & 4 \\ 
\addlinespace
Wordnet-scaled & I decanted water from the feeding bottle into the cup until \_ was empty. & feeding bottle & 4 \\ 
\bottomrule
\end{tabular}
\caption{Sample instances of \corpusname{} derived from an original WSC sentence, showcasing the different methods of sentence generation and perturbation.}
\label{tab:sample_instances}
\end{table*}

\subsection{The Dataset}

\looseness=-1 Table \ref{tab:dataset} outlines the construction and allocation process for our datasets, specifically \corpusname{}-small (S), \corpusname{}-medium (M) and \corpusname{}-large (L). The initial dataset, \corpusname{}-S, comprised 182 instances, all of which were adaptations induced by humans from an original set of 14 sentences.

Subsequently, we generated the \corpusname{}-M dataset, which was divided into three distinct subsets: `train', `val', and `test'. These subsets were created by perturbing the original sentences using ChatGPT, resulting in a total of 1,414 instances.

Finally, our most extensive dataset, \corpusname{}-L, was constructed by augmenting both the `train' and `val' subsets of \corpusname{}-M using Wordnet, leading to an overall count of 3,691 instances. The `test' subset was retained from the \corpusname{}-M `test' dataset and was generated through further perturbation of \corpusname{}-S sentences via ChatGPT. To illustrate the range of perturbations and their sources, we provide sample instances in Table \ref{tab:sample_instances} derived from an original WSC sentence.

\subsection{The Platform}
\label{sec:platform}
\looseness=-1 To foster collaborative development of \corpusname{}, we have developed an interactive platform, accessible at \url{https://evograd.com}. Here, global users can actively contribute to the dataset's evolution by modifying existing sentences.

In the \textit{Build dataset} page, users can select an original or perturbed sentence from a drop-down menu labeled \textit{Original Sentence}. They are then guided to input a modified version of this sentence, replacing the target pronoun with an underscore, in the \textit{New Sentence} field. Following the Winogrande format \cite{sakaguchi2020winogrande}, users also provide the two potential noun antecedents in the \textit{Option 1} and \textit{Option 2} fields, specifying the correct answer.

To enhance user engagement, our platform offers immediate feedback. Users can choose an LLM from a list - including BERT \cite{devlin-etal-2019-bert}, RoBERTa \cite{liu2019roberta}, and Albert \cite{lan2020albert}—and observe the model's live prediction. By clicking \textit{Submit}, this prediction is generated, and the newly provided data is incorporated into the dataset.

We prioritize transparency by allowing the dataset, stored as a CSV file, to be downloaded and inspected directly from the platform. To ensure the quality and appropriateness of the submissions, we manually validate all entries. Users are further supported with examples and guidelines. A glimpse of the platform's interface is depicted in Figure \ref{fig:interface}.

\begin{table*}[ht]
\small 
\begin{center}
\begin{tabular}{L{10cm}L{1.5cm}L{1.5cm}L{1cm}}
\toprule
\multicolumn{4}{p{15cm}}{\textbf{Original sentence}: Although she was being prosecuted, Monica was welcomed into the sanctuary of the church by Samantha because _ was a sinful criminal.} \\ \bottomrule 
\toprule

Perturbed Sentence & Prediction & True Label & Depth  \\
\midrule
    Although she was being prosecuted, Monica was welcomed into the sanctuary of the church by Samantha because _ was a guilty criminal. & Monica & Monica & 1 \ \textcolor{green}{\ding{51}} \\

\midrule
Although she was being prosecuted, Monica was welcomed into the sanctuary of the church by Samantha because _ was a compassionate person. & Samantha & Samantha & 2 \ \textcolor{green}{\ding{51}} \\

\midrule

Even though she was being prosecuted, Monica was guided into the safe haven of the church by Samantha because _ was a virtuous person.& Monica & Samantha & \textbf{5} \ \textcolor{red}{\ding{55}} \\

\midrule
While under prosecution, Monica was brought into the spiritual refuge of the church by Samantha because _ was a good-natured woman. & Monica & Samantha & \textbf{6} \textcolor{red}{\ding{55}} \\
    \midrule
    While being prosecuted, Monica was welcomed into the church's refuge by Samantha because _ was a law-abiding person. & Monica & Samantha & \textbf{5} \ \textcolor{red}{\ding{55}} \\
    
    \bottomrule

\end{tabular}
\caption{Sample of perturbations constructed from Eq.\ref{perturbation_equation} on a Winogrande example, with predictions corresponding to RoBERTa fine-tuned on Winogrande-XL. The model's incorrect predictions occur at depths 5,6 and 5, respectively, corresponding to the number of modified tokens from the original. 
Therefore, this sample of 5 perturbed instances has an average error depth (ED) of 5.333.}
\label{tab:running_example}
\end{center}
\end{table*}

\subsection{Error Depth}
\label{sec:metric}

Given our dataset construction methodology, we propose the \textit{error depth} (ED) metric to evaluate model stability. While accuracy is a widely used metric to gauge model performance on prediction tasks such as the WSC, it might not effectively capture a model's resilience against instances that progressively deviate from the original.

There are scenarios where models predict correctly but possibly for the wrong reasons. Sole reliance on accuracy can obscure these nuances. Ideally, a model should demonstrate stability against token substitutions. Although, in the context of the WSC, a token change can alter the answer label, a truly robust model should not be overly sensitive to such modifications.

The error depth metric quantifies a model's performance on sentences that increasingly diverge from a correctly understood original. Specifically, the error depth denotes the number of perturbations made to the original sentence before the model produces its first incorrect prediction.

For clarity, let's define the symbols:
\begin{itemize}
    \item \( s_{0} \): The original seed sentence.
    \item \( \text{label}(s) \): The true label of sentence \( s \).
    \item \( \text{pred}(s) \): The model's predicted label for sentence \( s \).
    \item \( n_{wrong} \): The number of incorrect predictions made by the model on perturbed versions of the original sentence.
\end{itemize}

With these definitions, the error depth (ED) is formulated as:
\vspace{-2mm}
\begin{equation} 
    \label{med_equation}
    \begin{aligned}
     \overline{ED} &\overset{def}= \frac{1}{n_{wrong}}\sum_{k}^{n_{wrong}}k \\
     &\text{if } \text{label}(s_{0}) = \text{pred}(s_{0}) \text{ and } \\
     &\text{label}(s_{k(i_1,...,i_k)}) \neq \text{pred}(s_{k(i_1,...,i_k)})
    \end{aligned}
\end{equation}

Refer to Table \ref{tab:running_example} for an application of the metric to perturbations of a sentence. In this demonstration, the model mispredicts three sentences: two after five perturbations and one after six. Thus, \( \overline{ED} = (5+5+6)/3 = 5.333 \). The error depth functions as an instance-level metric, assessing a model's stability for individual sentences. Averaging over all instances yields \( \overline{ED} \), which, when paired with accuracy, offers a comprehensive assessment of a model's performance on tasks like the WSC.

\subsection{Human Performance}

Three English-proficient annotators reviewed \corpusname{}-M Val and \corpusname{}-L Val, achieving mean accuracies of 95.2\% and 92.8\%, respectively. Importantly, they did not exhibit an average error depth, effectively handling perturbations to the full depth of the dataset. A high inter-annotator agreement was recorded with a Fleiss' Kappa of \(\kappa=0.914\).

\begin{table*}[ht]
    \centering
    \small
    \begin{tabular}{lllll}
        \toprule
        \textbf{Model} & \textbf{Tuning} & \textbf{Wino-valid} & \textbf{\corpusname{}-M-val} & \textbf{\corpusname{}-L-val} \\
        \midrule
        \multirow{5}{*}{BERT} & \corpusname{}-M & - & 60.4 (6.913) & - \\
        & \corpusname{}-L & - & - & 54.9 (6.867) \\
        & Wino & 62.75 & ----- (7.302) & ----- (7.258) \\
        & Wino + \corpusname{}-M & \textbf{63.06} & ----- (7.308) & - \\
        & Wino + \corpusname{}-L & 62.98 & - & ----- (7.232) \\
        \midrule
        \multirow{5}{*}{RoBERTa} & \corpusname{}-M & - & 58.4 (6.762) & - \\
        & \corpusname{}-L & - & - & 60.3 (6.727) \\
        & Wino & 76.09 & ----- (6.286) & 6.393 \\
        & Wino + \corpusname{}-M & 76.09 & ----- (6.286) & - \\
        & Wino + \corpusname{}-L & \textbf{76.64} & - & 6.652 \\
        \midrule
        \multirow{5}{*}{ALBERT} & \corpusname{}-M & - & 55.4 (6.989) & - \\
        & \corpusname{}-L & - & - & 57.2 (6.853) \\
        & Wino & 64.64 & ----- (7.971) & ----- (7.670) \\
        & Wino + \corpusname{}-M & 64.48 & ----- (8.000) & - \\
        & Wino + \corpusname{}-L & \textbf{64.64} & - & ----- (\underline{7.694}) \\
        \midrule
        \multirow{3}{*}{GPT-3*} & \corpusname{}-M & - & 59.41 (7.122) & - \\
        & \corpusname{}-L & - & - & 56.08 (6.753) \\
        \midrule
        \multirow{3}{*}{GPT-3.5*} & \corpusname{}-M & - & \textbf{67.33} (7.061) & - \\
        & \corpusname{}-L & - & - & \textbf{65.02} (7.245) \\
        \bottomrule
    \end{tabular}
    \caption{
    Accuracy (and error depth) results of models on Winogrande-valid and \corpusname{}-val sets after training on Winogrande-XL and/or \corpusname{}-train. \textbf{Bold} values represent the highest accuracy and \underline{underlined} values represent the highest error depth for each model in each dataset. A single dash (-) denotes that the model was not tuned on that specific dataset variant, hence was not tested. Dashed (-----) values indicate that accuracy was not tested due to potential contamination from \corpusname{}'s seed examples being taken from Winogrande, though error depth was still evaluated. Models marked with an asterisk (*) were evaluated using few-shot learning rather than fine-tuning.}
    \label{tab:results_accuracy}
\end{table*}

\section{Experiments and Results}
\subsection{Model Setup}
\label{sec:model_setup}

We evaluated three primary transformer-based models that are masked language models: BERT \cite{devlin-etal-2019-bert}, RoBERTa \cite{liu2019roberta}, and ALBERT \cite{lan2020albert}. These models have been recognized for their strong performance on the WSC and have led the benchmark results.  Each of the models were fine-tuned on the Winogrande-XL dataset \cite{sakaguchi2020winogrande}, which contains approximately 40,000 task instances and is designed to reduce potential annotation biases. 

Additionally, we evaluated two left-to-right language models, specifically GPT-3 (text-davinci-003) \cite{brown2020language} and GPT-3.5 (gpt-3.5-turbo-0613), on the Winogrande-XL and \corpusname{} datasets.

For BERT and RoBERTa, we first aimed to replicate top-performing models from existing literature. Using Huggingface's package \cite{wolf-etal-2020-transformers}, we achieved validation accuracies of 62.75\% for BERT-large-uncased and 76.09\% for RoBERTa-large. Although slightly below the reported accuracies in \cite{sakaguchi2020winogrande}, variations in hyperparameter tuning may account for the differences. A similar approach was taken for ALBERT-large-v2, with a resulting accuracy of 64.64\%.

Hyperparameters for BERT, RoBERTa, and ALBERT were selected from:
\begin{itemize}
    \item Learning rates: \(1e - 5\), \(3e - 5\), \(5e - 5\)
    \item Epochs: 3, 4, 5, 8
    \item Batch sizes: 8, 16
\end{itemize}

For training on \corpusname{}-train (both medium and large versions), given its resemblance but smaller size to Winogrande, we experimented with:
\begin{itemize}
    \item Learning rates: \(1e - 5\), \(3e - 5\), \(5e - 5\)
    \item Epochs: 1, 2, 4, 8
    \item Batch sizes: 8, 16, 32, 64
\end{itemize}

For evaluations using GPT-based models, we adopted a few-shot learning approach. Each instance was evaluated using an instruction-based prompt consisting of 30 random instances from the respective training set.

\begin{figure*}[h]
    \centering
    \includegraphics[width=\textwidth]{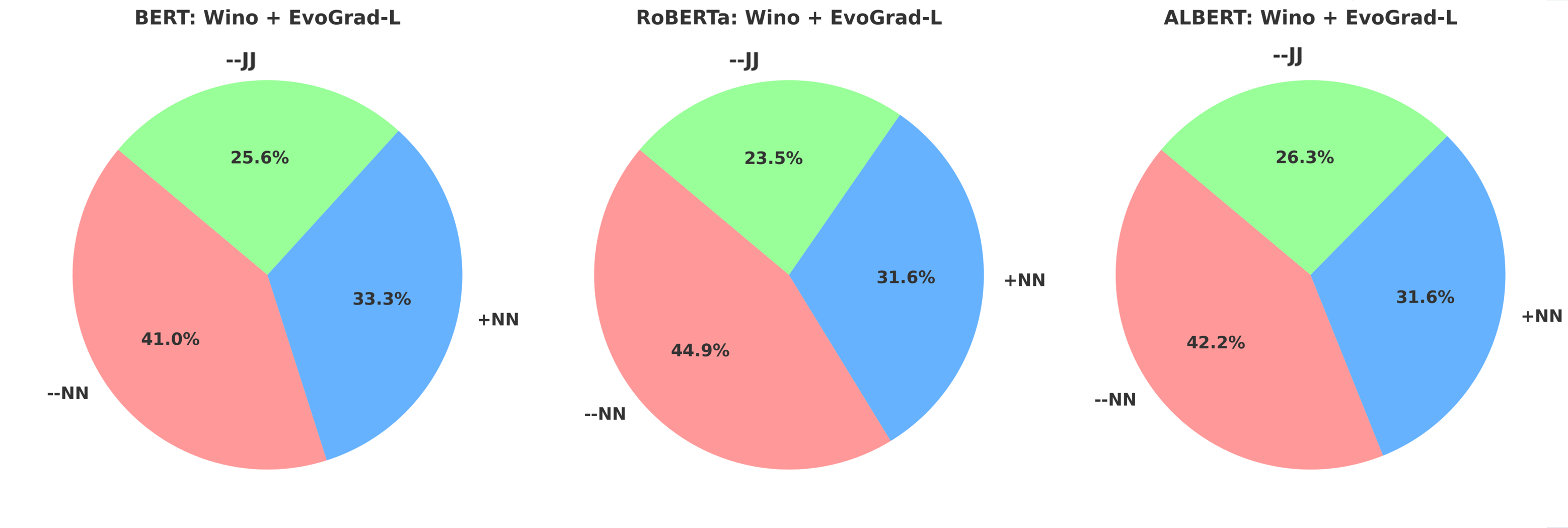}
    \vspace{-12mm}
    \caption{Distribution of the top three perturbations for models trained on Winogrande + Evograd-L. From left to right: BERT, RoBERTa, and ALBERT.The segments represent the relative frequency of each perturbation type: `--NN' (noun removal), `+NN' (noun addition), and either `--JJ' (adjective removal) or `--IN' (preposition removal).}
    \label{fig:pie_charts_colored}
\end{figure*}

\begin{table*}[ht]
    \centering
    \resizebox{\textwidth}{!}{
    \begin{tabular}{ |l|l|l|l|}
        \hline
        \textbf{Model} & \textbf{Trained on} & \textbf{\corpusname{}-M-val} & \textbf{\corpusname{}-L-val} \\
        \hline
        BERT & \corpusname{}-M & +NN (150), --NN (148), --JJ (105) & - \\
        & \corpusname{}-L & - & --NN (578), +NN (471), --JJ (342) \\
        &Wino & --NN (92), --JJ (62), +NN (61) & --NN (365), +NN (294), --JJ (228) \\
        & Wino + \corpusname{}-M & --NN (108), +NN (90), --JJ (78) & - \\
        & Wino + \corpusname{}-L & - & --NN (373), +NN (303), --JJ (233) \\
        \hline
        RoBERTa & \corpusname{}-M & --NN (170), +NN (146), --JJ (120) & - \\
        & \corpusname{}-L & - & --NN (494), +NN (416), --JJ (283) \\
        & Wino & --NN (17), +NN (12), --JJ (11) & --NN (76), +NN (54), --JJ (41) \\
        & Wino + \corpusname{}-M & --NN (17), +NN (12), --JJ (11) & - \\
         & Wino + \corpusname{}-L & - & --NN (61), +NN (43), --IN (32)  \\
        \hline
        ALBERT & \corpusname{}-M & --NN (189), +NN (161), --JJ (131) & - \\
        & \corpusname{}-L & - & --NN (542), +NN (479), --JJ (316) \\
        & Wino & --NN (92), --JJ (62), +NN (61) & --NN (272), +NN (208), --JJ (169) \\
        & Wino + \corpusname{}-M & --NN (92), --JJ (62), +NN (61) & -  \\
        & Wino + \corpusname{}-L & - & --NN (294), +NN (220), --JJ (183) \\
        \hline
        GPT-3 & \corpusname{}-M & --NN (173), +NN (144), --JJ (118) & - \\
        & \corpusname{}-L & - & --NN (505), +NN (448), --JJ (306) \\
        \hline
        GPT-3.5 & \corpusname{}-M & --NN (161), --JJ (115), +NN(111) & - \\
        & \corpusname{}-L & - & --NN (464), +NN (364), --JJ (290) \\
        \hline
        
    \end{tabular}}
    \caption{Top 3 perturbations and their count on incorrect predictions on \corpusname{}-val sets after fine-tuning on Winogrande-XL and \corpusname{}-train.}
    \label{tab:results_perturbations}
\end{table*}

\subsection{Results}
Our evaluation results, as shown in Tables \ref{tab:results_accuracy} and Figure \ref{fig:pie_charts_colored}, offer insight into model performance under different training conditions. We trained models exclusively on \corpusname{}-train, on Winogrande-XL (denoted as \textit{Wino}), or sequentially on both Winogrande and \corpusname{}-train (denoted as \textit{Wino + \corpusname{}}). This approach allowed us to understand how different training datasets influence model robustness and stability.

Table \ref{tab:results_accuracy} displays the models' accuracies on the Winogrande-valid dataset alongside their average error depth on the \corpusname{} datasets. The error depth indicates the perturbative distance at which a model starts to fail, providing insights into model stability. While accuracy is the main metric, error depth (shown in parentheses) gives a complementary view of model performance. Due to the potential overlap between \corpusname{} and Winogrande, we have omitted the accuracy scores for Winogrande-trained models in \corpusname{}. GPT-based models were only evaluated on \corpusname{} instances as they are evaluated through few-shot learning.

Figure \ref{fig:pie_charts_colored} visualizes the three most frequent perturbation types that lead to incorrect predictions by the models. Each perturbation is categorized by its effect on parts of speech. For instance, ``+NN (150)'' indicates a noun was added in 150 of the incorrect predictions. A comprehensive breakdown of the perturbation counts and their types, spanning all parts of speech observed, is provided in Table \ref{tab:results_perturbations}.

\section{Discussion}

\paragraph{Influence of \corpusname{} on Language Model Performance}

Table \ref{tab:results_accuracy} illustrates the varied impacts of \corpusname{} on Transformer models, leading to several key insights:

\begin{itemize}
\item BERT's improved performance post-\corpusname{} training underscores its ability to integrate the dataset's specific perturbations effectively. This adaptability implies that BERT may be particularly effective for tasks requiring deeper linguistic insight or sensitivity to subtle contextual changes.
\item RoBERTa consistently performs well both before and after training \corpusname{}, showcasing its robustness. However, its lower error depth compared to its accuracy points to a potential trade-off between performance and stability. This observation underscores the need to balance generalization with stability to perturbations.
\item The negligible change in ALBERT's performance across various training regimes raises questions regarding the model's saturation point and its alignment with the dataset. This warrants further investigation of the limits of adaptability for certain models.
\item While GPT-based models, especially GPT-3.5, demonstrate competitive performance, their error depths highlight challenges related to stability. This trend suggests that some of the newer models might prioritize adaptability at the expense of robustness.
\end{itemize}

Figure \ref{fig:pie_charts_colored} sheds light on the areas where language models are most vulnerable, particularly in handling noun and adjective modifications. Addressing these specific challenges is imperative for the enhancement of common-sense reasoning in future model iterations.

\paragraph{Robustness and Adaptability to New Tasks}
One of the challenges in deep learning is ensuring that the models remain adaptable and robust when exposed to new tasks or datasets. Whether through fine-tuning or few-shot learning, a model's ability to incorporate new information without significant detriment to its original capabilities is vital. In our experiments, the transformer models exhibited this adaptability, particularly when introduced to
\corpusname{}. For instance, when models were fine-tuned on 
\corpusname{}, their performance on the Winogrande validation set generally improved or remained consistent (Table \ref{tab:results_accuracy}), indicating that they did not lose their grasp of previously acquired knowledge. However, GPT-based models, through few-shot learning, demonstrated their versatility in quickly adapting to new tasks without the need for extensive retraining. These observations underscore the potential of current architectures in handling evolving datasets and tasks, highlighting their robustness in diverse learning scenarios. 

\vspace{-2mm}

\paragraph{Evolution and Community Involvement with \corpusname{}}

The current rendition of \corpusname{} represents only the first phase in a series of envisioned enhancements. As the platform matures, our goal is to achieve multiple cycles of data augmentation, model training, and fine-tuning, striving to foster a greater social impact in the AI domain. In making \corpusname{} accessible to a diverse audience, including those new to WSC-style challenges, we have incorporated clear prompts and guidelines, drawing inspiration from our initial work with the 182 instances in \corpusname{}-small.

Looking ahead, we are also planning to expand the platform to incorporate other foundational NLP tasks by integrating datasets such as OntoNotes 5.0 for Named Entity Recognition (NER) \cite{weischedel2012ontonotes}, Natural Questions (NQ) \cite{kwiatkowski2019natural} for Question Answering (QA), and the SemEval tasks for Sentiment Analysis, thereby broadening the scope and utility of \corpusname{}.

Recognizing the scale at which \corpusname{} could grow, we understand the crucial role of user-driven validation. While our dedicated team of in-house researchers currently curates the dataset to ensure its quality, we're eager to transition this role to our users in the near future. This strategy not only offloads the validation responsibility but also promises a more dynamic, participatory, and community-centric approach to refining LLMs.

\section{Conclusion}

In this work, we introduced \textbf{\corpusname{}}, a dynamic platform that extends the Winograd Schema Challenge with a human-and-model-in-the-loop methodology. The dataset, enriched through our platform, incorporates contributions from human experts, language model expansions, and lexical resource utilization. We also introduced the ``error depth" metric as a novel means to assess model stability in evolving tasks. While our evaluations showed potential benefits of using the augmented data from \corpusname{} across different training regimes, the disparity between human and machine performance on this task underlines its complexity and the ongoing challenges in enhancing common-sense reasoning in LLMs.




\section*{Ethics Statement}


We are presenting our publicly-accessible platform to those outside the scientific and crowd-sourcing communities. However, our platform is still limited to those in society who have access to a mobile device/personal computer and internet access; a large but underrepresented group of people in the world do not. We therefore use our platform as only a first step towards more inclusiveness, which we open to people outside the small community of science and crowd-sourcing, but wish to be involved in efforts that will include the underrepresented mentioned.

We also cannot assume that everyone's foremost priorities involve contributing towards such endeavours as ours -- many members of society are currently in turmoils of war, famine, or even indifference or aversion towards AI, which all may amount to their non-involvement in projects related to ours. Accordingly, the direction towards progress is best achieved \textit{outside the laboratory}; after all, if diversity and community involvement in the development of tasks such as ours is as correlated to positive results in AI, our efforts as researchers should also extend towards the education, well being, and thriving of members in society, without which our goal of a global task is never truly realized.

\section*{Acknowledgements}
This work was supported by the Natural Sciences
and Engineering Research Council of Canada (NSERC) and by the New Frontiers in Research Fund (NFRF).

\nocite{*}
\section{Bibliographical References}\label{sec:reference}

\bibliographystyle{lrec-coling2024-natbib}
\bibliography{custom}

\begin{thebibliography}{0}
\expandafter\ifx\csname natexlab\endcsname\relax\def\natexlab#1{#1}\fi

\end{thebibliography}


\begin{thebibliography}{85}
\expandafter\ifx\csname natexlab\endcsname\relax\def\natexlab#1{#1}\fi

\bibitem[{Abdou et~al.(2020)Abdou, Ravishankar, Barrett, Belinkov, Elliott, and
  S{\o}gaard}]{abdou-etal-2020-sensitivity}
Mostafa Abdou, Vinit Ravishankar, Maria Barrett, Yonatan Belinkov, Desmond
  Elliott, and Anders S{\o}gaard. 2020.
\newblock \href {https://doi.org/10.18653/v1/2020.acl-main.679} {The
  sensitivity of language models and humans to {W}inograd schema
  perturbations}.
\newblock In \emph{Proceedings of the 58th Annual Meeting of the Association
  for Computational Linguistics}, pages 7590--7604, Online. Association for
  Computational Linguistics.

\bibitem[{Aho and Ullman(1972)}]{Aho:72}
Alfred~V. Aho and Jeffrey~D. Ullman. 1972.
\newblock \emph{The Theory of Parsing, Translation and Compiling}, volume~1.
\newblock Prentice-Hall, Englewood Cliffs, NJ.

\bibitem[{{American Psychological Association}(1983)}]{APA:83}
{American Psychological Association}. 1983.
\newblock \emph{Publications Manual}.
\newblock American Psychological Association, Washington, DC.

\bibitem[{Anderson et~al.(2002)Anderson, Cobb, Korpela, Lebofsky, and
  Werthimer}]{anderson2002seti}
David~P Anderson, Jeff Cobb, Eric Korpela, Matt Lebofsky, and Dan Werthimer.
  2002.
\newblock Seti@ home: an experiment in public-resource computing.
\newblock \emph{Communications of the ACM}, 45(11):56--61.

\bibitem[{Ando and Zhang(2005)}]{Ando2005}
Rie~Kubota Ando and Tong Zhang. 2005.
\newblock \href {https://www.jmlr.org/papers/volume6/ando05a/ando05a.pdf} {A
  framework for learning predictive structures from multiple tasks and
  unlabeled data}.
\newblock \emph{Journal of Machine Learning Research}, 6:1817--1853.

\bibitem[{Andrew and Gao(2007)}]{andrew2007scalable}
Galen Andrew and Jianfeng Gao. 2007.
\newblock \href {https://dl.acm.org/doi/abs/10.1145/1273496.1273501} {Scalable
  training of {$L_1$}-regularized log-linear models}.
\newblock In \emph{Proceedings of the 24th International Conference on Machine
  Learning}, pages 33--40.

\bibitem[{Balasubramanian et~al.(2020)Balasubramanian, Jain, Jindal, Awasthi,
  and Sarawagi}]{balasubramanian-etal-2020-whats}
Sriram Balasubramanian, Naman Jain, Gaurav Jindal, Abhijeet Awasthi, and Sunita
  Sarawagi. 2020.
\newblock \href {https://doi.org/10.18653/v1/2020.repl4nlp-1.24} {What{'}s in a
  name? are {BERT} named entity representations just as good for any other
  name?}
\newblock In \emph{Proceedings of the 5th Workshop on Representation Learning
  for NLP}, pages 205--214, Online. Association for Computational Linguistics.

\bibitem[{Bergmanis et~al.(2017)Bergmanis, Kann, Sch{\"u}tze, and
  Goldwater}]{bergmanis-etal-2017-training}
Toms Bergmanis, Katharina Kann, Hinrich Sch{\"u}tze, and Sharon Goldwater.
  2017.
\newblock \href {https://doi.org/10.18653/v1/K17-2002} {Training data
  augmentation for low-resource morphological inflection}.
\newblock In \emph{Proceedings of the {C}o{NLL} {SIGMORPHON} 2017 Shared Task:
  Universal Morphological Reinflection}, pages 31--39, Vancouver. Association
  for Computational Linguistics.

\bibitem[{Bird and Klein(2009)}]{nltk}
Edward~Loper Bird, Steven and Ewan Klein. 2009.
\newblock \emph{Natural Language Processing with Python}.
\newblock O'Reilly Media Inc.

\bibitem[{Brabham(2013)}]{brabham2013crowdsourcing}
Daren~C. Brabham. 2013.
\newblock \emph{Crowdsourcing}.
\newblock MIT Press.

\bibitem[{Brown et~al.(2020)Brown, Mann, Ryder, Subbiah, Kaplan, Dhariwal,
  Neelakantan, Shyam, Sastry, Askell, Agarwal, Herbert-Voss, Krueger, Henighan,
  Child, Ramesh, Ziegler, Wu, Winter, Hesse, Chen, Sigler, Litwin, Gray, Chess,
  Clark, Berner, McCandlish, Radford, Sutskever, and
  Amodei}]{brown2020language}
Tom Brown, Benjamin Mann, Nick Ryder, Melanie Subbiah, Jared~D Kaplan, Prafulla
  Dhariwal, Arvind Neelakantan, Pranav Shyam, Girish Sastry, Amanda Askell,
  Sandhini Agarwal, Ariel Herbert-Voss, Gretchen Krueger, Tom Henighan, Rewon
  Child, Aditya Ramesh, Daniel Ziegler, Jeffrey Wu, Clemens Winter, Chris
  Hesse, Mark Chen, Eric Sigler, Mateusz Litwin, Scott Gray, Benjamin Chess,
  Jack Clark, Christopher Berner, Sam McCandlish, Alec Radford, Ilya Sutskever,
  and Dario Amodei. 2020.
\newblock \href
  {https://proceedings.neurips.cc/paper/2020/file/1457c0d6bfcb4967418bfb8ac142f64a-Paper.pdf}
  {Language models are few-shot learners}.
\newblock \emph{Advances in neural information processing systems},
  33:1877--1901.

\bibitem[{Chandra et~al.(1981)Chandra, Kozen, and Stockmeyer}]{Chandra:81}
Ashok~K. Chandra, Dexter~C. Kozen, and Larry~J. Stockmeyer. 1981.
\newblock \href {https://doi.org/10.1145/322234.322243} {Alternation}.
\newblock \emph{Journal of the Association for Computing Machinery},
  28(1):114--133.

\bibitem[{Chen et~al.(2019)Chen, D{'}Arcy, Liu, Fernandez, and
  Downey}]{chen-etal-2019-codah}
Michael Chen, Mike D{'}Arcy, Alisa Liu, Jared Fernandez, and Doug Downey. 2019.
\newblock \href {https://doi.org/10.18653/v1/W19-2008} {{CODAH}: An
  adversarially-authored question answering dataset for common sense}.
\newblock In \emph{Proceedings of the 3rd Workshop on Evaluating Vector Space
  Representations for {NLP}}, pages 63--69, Minneapolis, USA. Association for
  Computational Linguistics.

\bibitem[{Chen and Liu(2018)}]{chen2018lifelong}
Zhiyuan Chen and Bing Liu. 2018.
\newblock Lifelong machine learning.
\newblock \emph{Synthesis Lectures on Artificial Intelligence and Machine
  Learning}, 12(3):1--207.

\bibitem[{Cooley and Tukey(1965)}]{ct1965}
James~W. Cooley and John~W. Tukey. 1965.
\newblock \href
  {https://www.ams.org/journals/mcom/1965-19-090/S0025-5718-1965-0178586-1/S0025-5718-1965-0178586-1.pdf}
  {An algorithm for the machine calculation of complex {F}ourier series}.
\newblock \emph{Mathematics of Computation}, 19(90):297--301.

\bibitem[{Crowston(2012)}]{crowston2012amazon}
Kevin Crowston. 2012.
\newblock Amazon mechanical turk: A research tool for organizations and
  information systems scholars.
\newblock In \emph{Shaping the future of ict research. methods and approaches},
  pages 210--221. Springer.

\bibitem[{Dai et~al.(2023)Dai, Liu, Liao, Huang, Wu, Zhao, Liu, Liu, Li, Zhu
  et~al.}]{dai2023chataug}
Haixing Dai, Zhengliang Liu, Wenxiong Liao, Xiaoke Huang, Zihao Wu, Lin Zhao,
  Wei Liu, Ninghao Liu, Sheng Li, Dajiang Zhu, et~al. 2023.
\newblock Chataug: Leveraging chatgpt for text data augmentation.
\newblock \emph{arXiv preprint arXiv:2302.13007}.

\bibitem[{Devlin et~al.(2019)Devlin, Chang, Lee, and
  Toutanova}]{devlin-etal-2019-bert}
Jacob Devlin, Ming-Wei Chang, Kenton Lee, and Kristina Toutanova. 2019.
\newblock \href {https://doi.org/10.18653/v1/N19-1423} {{BERT}: Pre-training of
  deep bidirectional transformers for language understanding}.
\newblock In \emph{Proceedings of the 2019 Conference of the North {A}merican
  Chapter of the Association for Computational Linguistics: Human Language
  Technologies, Volume 1 (Long and Short Papers)}, pages 4171--4186,
  Minneapolis, Minnesota. Association for Computational Linguistics.

\bibitem[{Dinan et~al.(2019)Dinan, Humeau, Chintagunta, and
  Weston}]{dinan-etal-2019-build}
Emily Dinan, Samuel Humeau, Bharath Chintagunta, and Jason Weston. 2019.
\newblock \href {https://doi.org/10.18653/v1/D19-1461} {Build it break it fix
  it for dialogue safety: Robustness from adversarial human attack}.
\newblock In \emph{Proceedings of the 2019 Conference on Empirical Methods in
  Natural Language Processing and the 9th International Joint Conference on
  Natural Language Processing (EMNLP-IJCNLP)}, pages 4537--4546, Hong Kong,
  China. Association for Computational Linguistics.

\bibitem[{Elazar et~al.(2021{\natexlab{a}})Elazar, Zhang, Goldberg, and
  Roth}]{elazar-etal-2021-back}
Yanai Elazar, Hongming Zhang, Yoav Goldberg, and Dan Roth. 2021{\natexlab{a}}.
\newblock \href {https://doi.org/10.18653/v1/2021.emnlp-main.819} {Back to
  square one: Artifact detection, training and commonsense disentanglement in
  the {W}inograd schema}.
\newblock In \emph{Proceedings of the 2021 Conference on Empirical Methods in
  Natural Language Processing}, pages 10486--10500, Online and Punta Cana,
  Dominican Republic. Association for Computational Linguistics.

\bibitem[{Elazar et~al.(2021{\natexlab{b}})Elazar, Zhang, Goldberg, and
  Roth}]{elazar2021back}
Yanai Elazar, Hongming Zhang, Yoav Goldberg, and Dan Roth. 2021{\natexlab{b}}.
\newblock Back to square one: Artifact detection, training and commonsense
  disentanglement in the winograd schema.
\newblock In \emph{Proceedings of the 2021 Conference on Empirical Methods in
  Natural Language Processing}, pages 10486--10500.

\bibitem[{Emami et~al.(2020)Emami, Suleman, Trischler, and
  Cheung}]{emami2020analysis}
Ali Emami, Kaheer Suleman, Adam Trischler, and Jackie Chi~Kit Cheung. 2020.
\newblock An analysis of dataset overlap on winograd-style tasks.
\newblock In \emph{Proceedings of the 28th International Conference on
  Computational Linguistics}, pages 5855--5865.

\bibitem[{Emami et~al.(2019)Emami, Trichelair, Trischler, Suleman, Schulz, and
  Cheung}]{emami-etal-2019-knowref}
Ali Emami, Paul Trichelair, Adam Trischler, Kaheer Suleman, Hannes Schulz, and
  Jackie Chi~Kit Cheung. 2019.
\newblock \href {https://doi.org/10.18653/v1/P19-1386} {The {K}now{R}ef
  coreference corpus: Removing gender and number cues for difficult pronominal
  anaphora resolution}.
\newblock In \emph{Proceedings of the 57th Annual Meeting of the Association
  for Computational Linguistics}, pages 3952--3961, Florence, Italy.
  Association for Computational Linguistics.

\bibitem[{Fellbaum(2010)}]{fellbaum2010wordnet}
Christiane Fellbaum. 2010.
\newblock Wordnet.
\newblock In \emph{Theory and applications of ontology: computer applications},
  pages 231--243. Springer.

\bibitem[{Feng et~al.(2021)Feng, Gangal, Wei, Chandar, Vosoughi, Mitamura, and
  Hovy}]{feng-etal-2021-survey}
Steven~Y. Feng, Varun Gangal, Jason Wei, Sarath Chandar, Soroush Vosoughi,
  Teruko Mitamura, and Eduard Hovy. 2021.
\newblock \href {https://doi.org/10.18653/v1/2021.findings-acl.84} {A survey of
  data augmentation approaches for {NLP}}.
\newblock In \emph{Findings of the Association for Computational Linguistics:
  ACL-IJCNLP 2021}, pages 968--988, Online. Association for Computational
  Linguistics.

\bibitem[{French(1999)}]{french1999catastrophic}
Robert~M French. 1999.
\newblock Catastrophic forgetting in connectionist networks.
\newblock \emph{Trends in Cognitive Sciences}, 3(4):128--135.

\bibitem[{Gepperth and Hammer(2016)}]{gepperth2016incremental}
Alexander Gepperth and Barbara Hammer. 2016.
\newblock Incremental learning algorithms and applications.
\newblock In \emph{European symposium on artificial neural networks (ESANN)}.

\bibitem[{Gilardi et~al.(2023)Gilardi, Alizadeh, and
  Kubli}]{gilardi2023chatgpt}
Fabrizio Gilardi, Meysam Alizadeh, and Ma{\"e}l Kubli. 2023.
\newblock Chatgpt outperforms crowd-workers for text-annotation tasks.
\newblock \emph{arXiv preprint arXiv:2303.15056}.

\bibitem[{Gusfield(1997)}]{Gusfield:97}
Dan Gusfield. 1997.
\newblock \href
  {https://www.cambridge.org/core/books/algorithms-on-strings-trees-and-sequences/F0B095049C7E6EF5356F0A26686C20D3}
  {\emph{Algorithms on Strings, Trees and Sequences}}.
\newblock Cambridge University Press, Cambridge, UK.

\bibitem[{He et~al.(2021)He, Huang, Liu, and Zhu}]{he-etal-2021-winologic}
Weinan He, Canming Huang, Yongmei Liu, and Xiaodan Zhu. 2021.
\newblock \href {https://doi.org/10.18653/v1/2021.emnlp-main.307}
  {{W}ino{L}ogic: {A} zero-shot logic-based diagnostic dataset for {W}inograd
  {S}chema {C}hallenge}.
\newblock In \emph{Proceedings of the 2021 Conference on Empirical Methods in
  Natural Language Processing}, pages 3779--3789, Online and Punta Cana,
  Dominican Republic. Association for Computational Linguistics.

\bibitem[{Hossain and Kauranen(2015)}]{hossain2015crowdsourcing}
Mokter Hossain and Ilkka Kauranen. 2015.
\newblock Crowdsourcing: a comprehensive literature review.
\newblock \emph{Strategic Outsourcing: An International Journal}.

\bibitem[{Huang et~al.(2023)Huang, Kwak, and An}]{huang2023chatgpt}
Fan Huang, Haewoon Kwak, and Jisun An. 2023.
\newblock Is chatgpt better than human annotators? potential and limitations of
  chatgpt in explaining implicit hate speech.
\newblock In \emph{Companion Proceedings of the ACM Web Conference 2023}, pages
  294--297.

\bibitem[{Isaak and Michael(2019)}]{isaak2019winoflexi}
Nicos Isaak and Loizos Michael. 2019.
\newblock Winoflexi: A crowdsourcing platform for the development of winograd
  schemas.
\newblock In \emph{AI 2019: Advances in Artificial Intelligence}, pages
  289--302, Cham. Springer International Publishing.

\bibitem[{Isaak and Michael(2020)}]{isaak2020blending}
Nicos Isaak and Loizos Michael. 2020.
\newblock \href {https://doi.org/10.1007/978-3-030-71158-0_9} {Blending nlp and
  machine learning for the development of winograd schemas}.
\newblock In \emph{Agents and Artificial Intelligence: 12th International
  Conference, ICAART 2020, Valletta, Malta, February 22–24, 2020, Revised
  Selected Papers}, page 188–214, Berlin, Heidelberg. Springer-Verlag.

\bibitem[{Jia and Liang(2017)}]{jia-liang-2017-adversarial}
Robin Jia and Percy Liang. 2017.
\newblock \href {https://doi.org/10.18653/v1/D17-1215} {Adversarial examples
  for evaluating reading comprehension systems}.
\newblock In \emph{Proceedings of the 2017 Conference on Empirical Methods in
  Natural Language Processing}, pages 2021--2031, Copenhagen, Denmark.
  Association for Computational Linguistics.

\bibitem[{Kaushik et~al.(2020)Kaushik, Hovy, and Lipton}]{kaushik2020learning}
Divyansh Kaushik, Eduard Hovy, and Zachary~C Lipton. 2020.
\newblock Learning the difference that makes a difference with counterfactually
  augmented data.
\newblock \emph{International Conference on Learning Representations (ICLR)}.

\bibitem[{Kiela et~al.(2021)Kiela, Bartolo, Nie, Kaushik, Geiger, Wu, Vidgen,
  Prasad, Singh, Ringshia, Ma, Thrush, Riedel, Waseem, Stenetorp, Jia, Bansal,
  Potts, and Williams}]{kiela-etal-2021-dynabench}
Douwe Kiela, Max Bartolo, Yixin Nie, Divyansh Kaushik, Atticus Geiger,
  Zhengxuan Wu, Bertie Vidgen, Grusha Prasad, Amanpreet Singh, Pratik Ringshia,
  Zhiyi Ma, Tristan Thrush, Sebastian Riedel, Zeerak Waseem, Pontus Stenetorp,
  Robin Jia, Mohit Bansal, Christopher Potts, and Adina Williams. 2021.
\newblock \href {https://doi.org/10.18653/v1/2021.naacl-main.324} {Dynabench:
  Rethinking benchmarking in {NLP}}.
\newblock In \emph{Proceedings of the 2021 Conference of the North American
  Chapter of the Association for Computational Linguistics: Human Language
  Technologies}, pages 4110--4124, Online. Association for Computational
  Linguistics.

\bibitem[{Kirkpatrick et~al.(2017)Kirkpatrick, Pascanu, Rabinowitz, Veness,
  Desjardins, Rusu, Milan, Quan, Ramalho, Grabska-Barwinska, Hassabis, Clopath,
  Kumaran, and Hadsell}]{kirkpatrick2017overcoming}
James Kirkpatrick, Razvan Pascanu, Neil Rabinowitz, Joel Veness, Guillaume
  Desjardins, Andrei~A. Rusu, Kieran Milan, John Quan, Tiago Ramalho, Agnieszka
  Grabska-Barwinska, Demis Hassabis, Claudia Clopath, Dharshan Kumaran, and
  Raia Hadsell. 2017.
\newblock \href {https://doi.org/10.1073/pnas.1611835114} {Overcoming
  catastrophic forgetting in neural networks}.
\newblock \emph{Proceedings of the National Academy of Sciences},
  114(13):3521--3526.

\bibitem[{Kocijan et~al.(2019)Kocijan, Cretu, Camburu, Yordanov, and
  Lukasiewicz}]{kocijan-etal-2019-surprisingly}
Vid Kocijan, Ana-Maria Cretu, Oana-Maria Camburu, Yordan Yordanov, and Thomas
  Lukasiewicz. 2019.
\newblock \href {https://doi.org/10.18653/v1/P19-1478} {A surprisingly robust
  trick for the {W}inograd schema challenge}.
\newblock In \emph{Proceedings of the 57th Annual Meeting of the Association
  for Computational Linguistics}, pages 4837--4842, Florence, Italy.
  Association for Computational Linguistics.

\bibitem[{Kocijan et~al.(2022)Kocijan, Davis, Lukasiewicz, Marcus, and
  Morgenstern}]{kocijan2022defeat}
Vid Kocijan, Ernest Davis, Thomas Lukasiewicz, Gary Marcus, and Leora
  Morgenstern. 2022.
\newblock The defeat of the winograd schema challenge.
\newblock \emph{arXiv preprint arXiv:2201.02387}.

\bibitem[{Kratzwald and Feuerriegel(2019)}]{kratzwald2019learning}
Bernhard Kratzwald and Stefan Feuerriegel. 2019.
\newblock \href {https://doi.org/10.1145/3308558.3313661} {Learning from
  on-line user feedback in neural question answering on the web}.
\newblock In \emph{The World Wide Web Conference}, WWW '19, page 906–916, New
  York, NY, USA. Association for Computing Machinery.

\bibitem[{Kuzman et~al.(2023)Kuzman, Mozetic, and
  Ljube{\v{s}}ic}]{kuzman2023chatgpt}
Taja Kuzman, Igor Mozetic, and Nikola Ljube{\v{s}}ic. 2023.
\newblock Chatgpt: Beginning of an end of manual linguistic data annotation?
  use case of automatic genre identification.
\newblock \emph{ArXiv, abs/2303.03953}.

\bibitem[{Kwiatkowski et~al.(2019)Kwiatkowski, Palomaki, Redfield, Collins,
  Parikh, Alberti, Epstein, Polosukhin, Devlin, Lee
  et~al.}]{kwiatkowski2019natural}
Tom Kwiatkowski, Jennimaria Palomaki, Olivia Redfield, Michael Collins, Ankur
  Parikh, Chris Alberti, Danielle Epstein, Illia Polosukhin, Jacob Devlin,
  Kenton Lee, et~al. 2019.
\newblock Natural questions: a benchmark for question answering research.
\newblock \emph{Transactions of the Association for Computational Linguistics},
  7:453--466.

\bibitem[{Lan et~al.(2017)Lan, Qiu, He, and Xu}]{lan-etal-2017-continuously}
Wuwei Lan, Siyu Qiu, Hua He, and Wei Xu. 2017.
\newblock \href {https://doi.org/10.18653/v1/D17-1126} {A continuously growing
  dataset of sentential paraphrases}.
\newblock In \emph{Proceedings of the 2017 Conference on Empirical Methods in
  Natural Language Processing}, pages 1224--1234, Copenhagen, Denmark.
  Association for Computational Linguistics.

\bibitem[{Lan et~al.(2020)Lan, Chen, Goodman, Gimpel, Sharma, and
  Soricut}]{lan2020albert}
Zhenzhong Lan, Mingda Chen, Sebastian Goodman, Kevin Gimpel, Piyush Sharma, and
  Radu Soricut. 2020.
\newblock Albert: A lite bert for self-supervised learning of language
  representations.
\newblock In \emph{International Conference on Learning Representations}.

\bibitem[{Le~Bras et~al.(2020)Le~Bras, Swayamdipta, Bhagavatula, Zellers,
  Peters, Sabharwal, and Choi}]{lebras2020adversarial}
Ronan Le~Bras, Swabha Swayamdipta, Chandra Bhagavatula, Rowan Zellers, Matthew
  Peters, Ashish Sabharwal, and Yejin Choi. 2020.
\newblock Adversarial filters of dataset biases.
\newblock In \emph{International Conference on Machine Learning}, pages
  1078--1088. PMLR.

\bibitem[{Levesque et~al.(2011)Levesque, Davis, and
  Morgenstern}]{levesque2011winograd}
Hector Levesque, Ernest Davis, and Leora Morgenstern. 2011.
\newblock The winograd schema challenge.
\newblock In \emph{AAAI Spring Symposium: Logical Formalizations of Commonsense
  Reasoning}.

\bibitem[{Lin et~al.(2020{\natexlab{a}})Lin, Lee, Khanna, and
  Ren}]{lin-etal-2020-birds}
Bill~Yuchen Lin, Seyeon Lee, Rahul Khanna, and Xiang Ren. 2020{\natexlab{a}}.
\newblock \href {https://doi.org/10.18653/v1/2020.emnlp-main.557} {{B}irds have
  four legs?! {N}umer{S}ense: {P}robing {N}umerical {C}ommonsense {K}nowledge
  of {P}re-{T}rained {L}anguage {M}odels}.
\newblock In \emph{Proceedings of the 2020 Conference on Empirical Methods in
  Natural Language Processing (EMNLP)}, pages 6862--6868, Online. Association
  for Computational Linguistics.

\bibitem[{Lin et~al.(2020{\natexlab{b}})Lin, Zhou, Shen, Zhou, Bhagavatula,
  Choi, and Ren}]{lin-etal-2020-commongen}
Bill~Yuchen Lin, Wangchunshu Zhou, Ming Shen, Pei Zhou, Chandra Bhagavatula,
  Yejin Choi, and Xiang Ren. 2020{\natexlab{b}}.
\newblock \href {https://doi.org/10.18653/v1/2020.findings-emnlp.165}
  {{C}ommon{G}en: A constrained text generation challenge for generative
  commonsense reasoning}.
\newblock In \emph{Findings of the Association for Computational Linguistics:
  EMNLP 2020}, pages 1823--1840, Online. Association for Computational
  Linguistics.

\bibitem[{Liu et~al.(2019)Liu, Ott, Goyal, Du, Joshi, Chen, Levy, Lewis,
  Zettlemoyer, and Stoyanov}]{liu2019roberta}
Yinhan Liu, Myle Ott, Naman Goyal, Jingfei Du, Mandar Joshi, Danqi Chen, Omer
  Levy, Mike Lewis, Luke Zettlemoyer, and Veselin Stoyanov. 2019.
\newblock Roberta: A robustly optimized bert pretraining approach.
\newblock \emph{arXiv preprint arXiv:1907.11692}.

\bibitem[{Lov{\'o}n-Melgarejo et~al.(2021)Lov{\'o}n-Melgarejo, Soulier,
  Pinel-Sauvagnat, and Tamine}]{lovon2021studying}
Jes{\'u}s Lov{\'o}n-Melgarejo, Laure Soulier, Karen Pinel-Sauvagnat, and Lynda
  Tamine. 2021.
\newblock Studying catastrophic forgetting in neural ranking models.
\newblock In \emph{European Conference on Information Retrieval}, pages
  375--390. Springer.

\bibitem[{Lu et~al.(2022)Lu, Yang, Namee, and Zhang}]{lu-etal-2022-rationale}
Jinghui Lu, Linyi Yang, Brian Namee, and Yue Zhang. 2022.
\newblock \href {https://doi.org/10.18653/v1/2022.acl-long.481} {A
  rationale-centric framework for human-in-the-loop machine learning}.
\newblock In \emph{Proceedings of the 60th Annual Meeting of the Association
  for Computational Linguistics (Volume 1: Long Papers)}, pages 6986--6996,
  Dublin, Ireland. Association for Computational Linguistics.

\bibitem[{Lu et~al.(2020)Lu, Mardziel, Wu, Amancharla, and
  Datta}]{lu2020gender}
Kaiji Lu, Piotr Mardziel, Fangjing Wu, Preetam Amancharla, and Anupam Datta.
  2020.
\newblock Gender bias in neural natural language processing.
\newblock In \emph{Logic, Language, and Security}, pages 189--202. Springer.

\bibitem[{Melo et~al.(2019)Melo, Imaizumi, and Cozman}]{melo2019winograd}
Gabriela Melo, Vinicius Imaizumi, and Fábio Cozman. 2019.
\newblock \href {https://doi.org/10.5753/eniac.2019.9334} {Winograd schemas in
  portuguese}.
\newblock In \emph{Proceedings of 16th National Meeting on Artificial and
  Computational Intelligence}, pages 787--798, Porto Alegre, RS, Brasil. SBC.

\bibitem[{Mitchell et~al.(2015)Mitchell, Cohen, Hruschka, Talukdar, Betteridge,
  Carlson, Dalvi~Mishra, Gardner, Kisiel, Krishnamurthy, Lao, Mazaitis,
  Mohamed, Nakashole, Platanios, Ritter, Samadi, Settles, Wang, Wijaya, Gupta,
  Chen, Saparov, Greaves, and Welling}]{mitchell2015never}
Tom Mitchell, William Cohen, Estevam Hruschka, Partha Talukdar, Justin
  Betteridge, Andrew Carlson, Bhavana Dalvi~Mishra, Matthew Gardner, Bryan
  Kisiel, Jayant Krishnamurthy, Ni~Lao, Kathryn Mazaitis, Thahir Mohamed, Ndapa
  Nakashole, Emmanouil Platanios, Alan Ritter, Mehdi Samadi, Burr Settles,
  Richard Wang, Derry Wijaya, Abhinav Gupta, Xinlei Chen, Abulhair Saparov,
  Malcolm Greaves, and Joel Welling. 2015.
\newblock \href {https://doi.org/10.1609/aaai.v29i1.9498} {Never-ending
  learning}.
\newblock \emph{Proceedings of the AAAI Conference on Artificial Intelligence},
  29(1).

\bibitem[{Morgenstern et~al.(2016)Morgenstern, Davis, and
  Ortiz}]{morgenstern2016planning}
Leora Morgenstern, Ernest Davis, and Charles~L. Ortiz. 2016.
\newblock \href {https://doi.org/10.1609/aimag.v37i1.2639} {Planning,
  executing, and evaluating the winograd schema challenge}.
\newblock \emph{AI Magazine}, 37(1):50--54.

\bibitem[{Nie et~al.(2020)Nie, Williams, Dinan, Bansal, Weston, and
  Kiela}]{nie-etal-2020-adversarial}
Yixin Nie, Adina Williams, Emily Dinan, Mohit Bansal, Jason Weston, and Douwe
  Kiela. 2020.
\newblock \href {https://doi.org/10.18653/v1/2020.acl-main.441} {Adversarial
  {NLI}: A new benchmark for natural language understanding}.
\newblock In \emph{Proceedings of the 58th Annual Meeting of the Association
  for Computational Linguistics}, pages 4885--4901, Online. Association for
  Computational Linguistics.

\bibitem[{Rasooli and Tetreault(2015)}]{rasooli-tetrault-2015}
Mohammad~Sadegh Rasooli and Joel~R. Tetreault. 2015.
\newblock \href {http://arxiv.org/abs/1503.06733} {Yara parser: {A} fast and
  accurate dependency parser}.
\newblock \emph{Computing Research Repository}, arXiv:1503.06733.
\newblock Version 2.

\bibitem[{Rudinger et~al.(2018)Rudinger, Naradowsky, Leonard, and
  Van~Durme}]{rudinger-etal-2018-gender}
Rachel Rudinger, Jason Naradowsky, Brian Leonard, and Benjamin Van~Durme. 2018.
\newblock \href {https://doi.org/10.18653/v1/N18-2002} {Gender bias in
  coreference resolution}.
\newblock In \emph{Proceedings of the 2018 Conference of the North {A}merican
  Chapter of the Association for Computational Linguistics: Human Language
  Technologies, Volume 2 (Short Papers)}, pages 8--14, New Orleans, Louisiana.
  Association for Computational Linguistics.

\bibitem[{Sakaguchi et~al.(2020)Sakaguchi, Le~Bras, Bhagavatula, and
  Choi}]{sakaguchi2020winogrande}
Keisuke Sakaguchi, Ronan Le~Bras, Chandra Bhagavatula, and Yejin Choi. 2020.
\newblock Winogrande: An adversarial winograd schema challenge at scale.
\newblock In \emph{Proceedings of the AAAI Conference on Artificial
  Intelligence}, volume~34, pages 8732--8740.

\bibitem[{Shi et~al.(2021)Shi, Livescu, and
  Gimpel}]{shi-etal-2021-substructure}
Haoyue Shi, Karen Livescu, and Kevin Gimpel. 2021.
\newblock \href {https://doi.org/10.18653/v1/2021.findings-acl.307}
  {Substructure substitution: Structured data augmentation for {NLP}}.
\newblock In \emph{Findings of the Association for Computational Linguistics:
  ACL-IJCNLP 2021}, pages 3494--3508, Online. Association for Computational
  Linguistics.

\bibitem[{Shi et~al.(2018)Shi, Mao, Xiao, Jiang, and
  Sun}]{shi-etal-2018-learning}
Haoyue Shi, Jiayuan Mao, Tete Xiao, Yuning Jiang, and Jian Sun. 2018.
\newblock \href {https://aclanthology.org/C18-1315} {Learning visually-grounded
  semantics from contrastive adversarial samples}.
\newblock In \emph{Proceedings of the 27th International Conference on
  Computational Linguistics}, pages 3715--3727, Santa Fe, New Mexico, USA.
  Association for Computational Linguistics.

\bibitem[{Silver et~al.(2013)Silver, Yang, and Li}]{silver2013lifelong}
Daniel~L Silver, Qiang Yang, and Lianghao Li. 2013.
\newblock Lifelong machine learning systems: Beyond learning algorithms.
\newblock In \emph{2013 AAAI spring symposium series}.

\bibitem[{Trichelair et~al.(2018)Trichelair, Emami, Cheung, Trischler, Suleman,
  and Diaz}]{trichelair2018on}
Paul Trichelair, Ali Emami, Jackie Chi~Kit Cheung, Adam Trischler, Kaheer
  Suleman, and Fernando Diaz. 2018.
\newblock \href
  {https://www.microsoft.com/en-us/research/publication/on-the-evaluation-of-common-sense-reasoning-in-natural-language-understanding/}
  {On the evaluation of common-sense reasoning in natural language
  understanding}.
\newblock In \emph{Critiquing and Correcting Trends in Machine Learning NeurIPS
  2018 Workshop}.

\bibitem[{Trichelair et~al.(2019)Trichelair, Emami, Trischler, Suleman, and
  Cheung}]{trichelair-etal-2019-reasonable}
Paul Trichelair, Ali Emami, Adam Trischler, Kaheer Suleman, and Jackie Chi~Kit
  Cheung. 2019.
\newblock \href {https://doi.org/10.18653/v1/D19-1335} {How reasonable are
  common-sense reasoning tasks: A case-study on the {W}inograd schema challenge
  and {SWAG}}.
\newblock In \emph{Proceedings of the 2019 Conference on Empirical Methods in
  Natural Language Processing and the 9th International Joint Conference on
  Natural Language Processing (EMNLP-IJCNLP)}, pages 3382--3387, Hong Kong,
  China. Association for Computational Linguistics.

\bibitem[{Turing(1950)}]{turing1950computing}
Alan~M. Turing. 1950.
\newblock Computing machinery and intelligence.
\newblock \emph{Mind}.

\bibitem[{Vaswani et~al.(2017)Vaswani, Shazeer, Parmar, Uszkoreit, Jones,
  Gomez, Kaiser, and Polosukhin}]{vaswani2017attention}
Ashish Vaswani, Noam Shazeer, Niki Parmar, Jakob Uszkoreit, Llion Jones,
  Aidan~N Gomez, \L~ukasz Kaiser, and Illia Polosukhin. 2017.
\newblock \href
  {https://proceedings.neurips.cc/paper/2017/file/3f5ee243547dee91fbd053c1c4a845aa-Paper.pdf}
  {Attention is all you need}.
\newblock In \emph{Advances in Neural Information Processing Systems},
  volume~30. Curran Associates, Inc.

\bibitem[{Wallace et~al.(2019)Wallace, Rodriguez, Feng, Yamada, and
  Boyd-Graber}]{wallace-etal-2019-trick}
Eric Wallace, Pedro Rodriguez, Shi Feng, Ikuya Yamada, and Jordan Boyd-Graber.
  2019.
\newblock \href {https://doi.org/10.1162/tacl_a_00279} {Trick me if you can:
  Human-in-the-loop generation of adversarial examples for question answering}.
\newblock \emph{Transactions of the Association for Computational Linguistics},
  7:387--401.

\bibitem[{Wang et~al.(2018)Wang, Singh, Michael, Hill, Levy, and
  Bowman}]{wang-etal-2018-glue}
Alex Wang, Amanpreet Singh, Julian Michael, Felix Hill, Omer Levy, and Samuel
  Bowman. 2018.
\newblock \href {https://doi.org/10.18653/v1/W18-5446} {{GLUE}: A multi-task
  benchmark and analysis platform for natural language understanding}.
\newblock In \emph{Proceedings of the 2018 {EMNLP} Workshop {B}lackbox{NLP}:
  Analyzing and Interpreting Neural Networks for {NLP}}, pages 353--355,
  Brussels, Belgium. Association for Computational Linguistics.

\bibitem[{Wang and Yang(2015)}]{wang-yang-2015-thats}
William~Yang Wang and Diyi Yang. 2015.
\newblock \href {https://doi.org/10.18653/v1/D15-1306} {That{'}s so
  annoying!!!: A lexical and frame-semantic embedding based data augmentation
  approach to automatic categorization of annoying behaviors using {\#}petpeeve
  tweets}.
\newblock In \emph{Proceedings of the 2015 Conference on Empirical Methods in
  Natural Language Processing}, pages 2557--2563, Lisbon, Portugal. Association
  for Computational Linguistics.

\bibitem[{Weischedel et~al.(2012)Weischedel, Pradhan, Ramshaw, Kaufman,
  Franchini, El-Bachouti, Xue, Palmer, Hwang, Bonial
  et~al.}]{weischedel2012ontonotes}
R~Weischedel, S~Pradhan, L~Ramshaw, J~Kaufman, M~Franchini, M~El-Bachouti,
  N~Xue, M~Palmer, JD~Hwang, C~Bonial, et~al. 2012.
\newblock Ontonotes release 5.0. linguistic data consortium.
\newblock Technical report, Philadelphia, Technical Report.

\bibitem[{Winograd(1972)}]{winograd1972}
Terry Winograd. 1972.
\newblock \href {https://doi.org/https://doi.org/10.1016/0010-0285(72)90002-3}
  {Understanding natural language}.
\newblock \emph{Cognitive Psychology}, 3(1):1--191.

\bibitem[{Wolf et~al.(2020)Wolf, Debut, Sanh, Chaumond, Delangue, Moi, Cistac,
  Rault, Louf, Funtowicz, Davison, Shleifer, von Platen, Ma, Jernite, Plu, Xu,
  Le~Scao, Gugger, Drame, Lhoest, and Rush}]{wolf-etal-2020-transformers}
Thomas Wolf, Lysandre Debut, Victor Sanh, Julien Chaumond, Clement Delangue,
  Anthony Moi, Pierric Cistac, Tim Rault, Remi Louf, Morgan Funtowicz, Joe
  Davison, Sam Shleifer, Patrick von Platen, Clara Ma, Yacine Jernite, Julien
  Plu, Canwen Xu, Teven Le~Scao, Sylvain Gugger, Mariama Drame, Quentin Lhoest,
  and Alexander Rush. 2020.
\newblock \href {https://doi.org/10.18653/v1/2020.emnlp-demos.6} {Transformers:
  State-of-the-art natural language processing}.
\newblock In \emph{Proceedings of the 2020 Conference on Empirical Methods in
  Natural Language Processing: System Demonstrations}, pages 38--45, Online.
  Association for Computational Linguistics.

\bibitem[{Xu et~al.(2021)Xu, Ju, Li, Boureau, Weston, and
  Dinan}]{xu-etal-2021-bot}
Jing Xu, Da~Ju, Margaret Li, Y-Lan Boureau, Jason Weston, and Emily Dinan.
  2021.
\newblock \href {https://doi.org/10.18653/v1/2021.naacl-main.235}
  {Bot-adversarial dialogue for safe conversational agents}.
\newblock In \emph{Proceedings of the 2021 Conference of the North American
  Chapter of the Association for Computational Linguistics: Human Language
  Technologies}, pages 2950--2968, Online. Association for Computational
  Linguistics.

\bibitem[{Xu et~al.(2016)Xu, Jia, Mou, Li, Chen, Lu, and
  Jin}]{xu-etal-2016-improved}
Yan Xu, Ran Jia, Lili Mou, Ge~Li, Yunchuan Chen, Yangyang Lu, and Zhi Jin.
  2016.
\newblock \href {https://aclanthology.org/C16-1138} {Improved relation
  classification by deep recurrent neural networks with data augmentation}.
\newblock In \emph{Proceedings of {COLING} 2016, the 26th International
  Conference on Computational Linguistics: Technical Papers}, pages 1461--1470,
  Osaka, Japan. The COLING 2016 Organizing Committee.

\bibitem[{Yang et~al.(2018)Yang, Qi, Zhang, Bengio, Cohen, Salakhutdinov, and
  Manning}]{yang-etal-2018-hotpotqa}
Zhilin Yang, Peng Qi, Saizheng Zhang, Yoshua Bengio, William Cohen, Ruslan
  Salakhutdinov, and Christopher~D. Manning. 2018.
\newblock \href {https://doi.org/10.18653/v1/D18-1259} {{H}otpot{QA}: A dataset
  for diverse, explainable multi-hop question answering}.
\newblock In \emph{Proceedings of the 2018 Conference on Empirical Methods in
  Natural Language Processing}, pages 2369--2380, Brussels, Belgium.
  Association for Computational Linguistics.

\bibitem[{Zahraei and Emami(2024)}]{zahraei2024wsc+}
Pardis~Sadat Zahraei and Ali Emami. 2024.
\newblock Wsc+: Enhancing the winograd schema challenge using tree-of-experts.
\newblock \emph{arXiv preprint arXiv:2401.17703}.

\bibitem[{Zellers et~al.(2018)Zellers, Bisk, Schwartz, and
  Choi}]{zellers-etal-2018-swag}
Rowan Zellers, Yonatan Bisk, Roy Schwartz, and Yejin Choi. 2018.
\newblock \href {https://doi.org/10.18653/v1/D18-1009} {{SWAG}: A large-scale
  adversarial dataset for grounded commonsense inference}.
\newblock In \emph{Proceedings of the 2018 Conference on Empirical Methods in
  Natural Language Processing}, pages 93--104, Brussels, Belgium. Association
  for Computational Linguistics.

\bibitem[{Zellers et~al.(2019)Zellers, Holtzman, Bisk, Farhadi, and
  Choi}]{zellers-etal-2019-hellaswag}
Rowan Zellers, Ari Holtzman, Yonatan Bisk, Ali Farhadi, and Yejin Choi. 2019.
\newblock \href {https://doi.org/10.18653/v1/P19-1472} {{H}ella{S}wag: Can a
  machine really finish your sentence?}
\newblock In \emph{Proceedings of the 57th Annual Meeting of the Association
  for Computational Linguistics}, pages 4791--4800, Florence, Italy.
  Association for Computational Linguistics.

\bibitem[{Zhang et~al.(2020)Zhang, Zhao, and Song}]{zhang-etal-2020-winowhy}
Hongming Zhang, Xinran Zhao, and Yangqiu Song. 2020.
\newblock \href {https://doi.org/10.18653/v1/2020.acl-main.508} {{W}ino{W}hy: A
  deep diagnosis of essential commonsense knowledge for answering {W}inograd
  schema challenge}.
\newblock In \emph{Proceedings of the 58th Annual Meeting of the Association
  for Computational Linguistics}, pages 5736--5745, Online. Association for
  Computational Linguistics.

\bibitem[{Zhang et~al.(2018)Zhang, Cisse, Dauphin, and
  Lopez-Paz}]{zhang2018mixup}
Hongyi Zhang, Moustapha Cisse, Yann~N Dauphin, and David Lopez-Paz. 2018.
\newblock mixup: Beyond empirical risk minimization.
\newblock In \emph{International Conference on Learning Representations}.

\bibitem[{Zhang et~al.(2015)Zhang, Zhao, and LeCun}]{zhang2015character}
Xiang Zhang, Junbo Zhao, and Yann LeCun. 2015.
\newblock \href
  {https://proceedings.neurips.cc/paper/2015/file/250cf8b51c773f3f8dc8b4be867a9a02-Paper.pdf}
  {Character-level convolutional networks for text classification}.
\newblock In \emph{Advances in Neural Information Processing Systems},
  volume~28. Curran Associates, Inc.

\bibitem[{Zhao et~al.(2018)Zhao, Wang, Yatskar, Ordonez, and
  Chang}]{zhao-etal-2018-gender}
Jieyu Zhao, Tianlu Wang, Mark Yatskar, Vicente Ordonez, and Kai-Wei Chang.
  2018.
\newblock \href {https://doi.org/10.18653/v1/N18-2003} {Gender bias in
  coreference resolution: Evaluation and debiasing methods}.
\newblock In \emph{Proceedings of the 2018 Conference of the North {A}merican
  Chapter of the Association for Computational Linguistics: Human Language
  Technologies, Volume 2 (Short Papers)}, pages 15--20, New Orleans, Louisiana.
  Association for Computational Linguistics.

\bibitem[{Zhu et~al.(2023)Zhu, Zhang, Haq, Hui, and Tyson}]{zhu2023can}
Yiming Zhu, Peixian Zhang, Ehsan-Ul Haq, Pan Hui, and Gareth Tyson. 2023.
\newblock Can chatgpt reproduce human-generated labels? a study of social
  computing tasks.
\newblock \emph{arXiv preprint arXiv:2304.10145}.

\bibitem[{Zmigrod et~al.(2019)Zmigrod, Mielke, Wallach, and
  Cotterell}]{zmigrod-etal-2019-counterfactual}
Ran Zmigrod, Sabrina~J. Mielke, Hanna Wallach, and Ryan Cotterell. 2019.
\newblock \href {https://doi.org/10.18653/v1/P19-1161} {Counterfactual data
  augmentation for mitigating gender stereotypes in languages with rich
  morphology}.
\newblock In \emph{Proceedings of the 57th Annual Meeting of the Association
  for Computational Linguistics}, pages 1651--1661, Florence, Italy.
  Association for Computational Linguistics.

\end{thebibliography}

\onecolumn
\appendix
\section{Appendix}
\label{sec:appendix}

\subsection{ChatGPT Dialogue During Scaling Process}
\label{sec:chatgpt_dialogue}
\begin{description}

\item[Author:] 
You are an expert linguist, human annotator and are familiar with the Winograd Schema Challenge. Consider the sentence: Although they ran at about the same speed, Sue beat Sally because _ had such a good start.  This sentence is in the format of a Winograd schema, where two antecedents, in this case "Sue" and "Sally" are competing for reference in the position, _. Common sense allows us to conclude that one of the two antecedents is most likely, in this case, Sue. 

I am intending on building a dataset composed of modifications to the above sentence. Based on this original sentence, I need you to generate 100 new sentences that are modified by either adding, replacing, or substituting words in the original sentence. They should be organized in of the following format, copied from a csv file :

index	sentence	option1	option2	answer	distance

where the original sentence has the following form:
0	Although they ran at about the same speed, Sue beat Sally because _ had such a good start.	Sue	Sally	1	0

Here, index refers to the index number for the sentence, sentence refers to the sentence itself (notice the necessity of the _ as the position for one of the two antecedents), option1 refers to one antecedent, option2 refers to the other, answer refers to the correct antecedent that belongs in _ -- "1" corresponds to option1, and "2" corresponds to option2. Notice, for this original sentence, the answer is "1" because having a good start is a likely explanation for someone (i.e., Sue) beating another in a race. Finally, "distance" corresponds to the number of modifications made on the original sentence (whose distance is, by definition, 0). It is equivalent to the "token edit distance". 

I want you to create a diverse range of new entries, that vary in terms of the modifications, the total distance, as well as the correct answer.

For example, the sentence "Although they sprinted at about the same speed, Sue beat Sally because _ had such a good start." would have an answer of 1 and distance of 1 (because ran was changed to sprinted, but the correct answer still remains option1, which in this case is Sue), and the sentence "Although they sprinted at about the same speed, Sue beat Sally although _ had such a good start." would have the answer of 2 and distance of 2, because the correct answer is now "Sally", and there were two word modifications -- the second one here was "although" instead of "because", which flips the answer.  Notice that with the right change in words, the answer can flip -- I want to encourage you to find such minimal number of changes when you're creating new entries. You are also allowed to change the words in the options too (e.g., Sue and Sally). I would like the dataset to have a similar number of answer=1 entries than answer=2 entries.

Please generate the first 10, formatted as a table here. First, you may ask any clarification questions to make sure you understand the task.

\item[ChatGPT:] Thank you for the instructions. Here are the first 10 entries: [...]
 
\item[Author:] Excellent work. Please generate 30 more, but try to keep the distance lower. Don't forget you can change the names too!

\item[ChatGPT:] Sure! Here are the next 30 entries with smaller distances and changes in names: [...]

\item[Author:] These sentences didn't actually have the distances you claimed they had. These should correspond to token edit distances. Please be careful when generating sentences to actually correspond to distances -- it's okay if the distance is a bit higher. I just want you to be accurate. You also forgot to change the names (as possible modifications) at one point. Please generate 10 for me to see if it's okay. 

\item[ChatGPT:] I apologize for the inaccuracies in the distances and missing name changes in the previous examples. [...] Here are the next 10 sentences: [...]

\item[Author:] These were much better. Please generate 30 more. Pay attention to not miscalculate the distance.

\item[ChatGPT:] I appreciate your feedback. Here are the next 30 sentences, with the correct token edit distances and a careful attention to detail: [...]

\item[Author:] Great job. Please generate 30 more. Again, the token distance should be correctly calculated. For example, in your sentence:

Even though they raced at the same speed, Sue beat Sally although _ had a powerful start. 

You said the distance is 3. It's actually 5 here, because Although was dropped and replaced with Even (1), then "though" was added (2), then raced was replaced by ran (3), then about was dropped (4), then good was replaced by powerful (5).

\item[ChatGPT:] I understand your concern and I apologize for the mistake. I'll ensure more accuracy in calculating the token edit distance. Here are the next 30 sentences: [...]

\item[Author:] Great job. Please generate 20 more. Remember that changing the names is valid but counts as a modification (increments distance).

\item[ChatGPT:] Thank you for your feedback. Here are the next 20 sentences, this time including changes in names: [...]

\end{description}

\bibliographystylelanguageresource{lrec-coling2024-natbib}
\bibliographylanguageresource{languageresource}

\end{document}